\RequirePackage{fix-cm}
\documentclass[twocolumn]{svjour3}
\smartqed{}
\usepackage{etoolbox}
\usepackage[utf8]{inputenc}
\usepackage[T1]{fontenc}
\usepackage[UKenglish]{babel}
\usepackage{amsmath}
\usepackage{amssymb}
\usepackage{mathtools}
\usepackage{array}
\usepackage{multirow}
\usepackage{booktabs}
\usepackage{graphicx}
\usepackage{url}
\usepackage[sort]{cite}
\usepackage{pgfplots}
\pgfplotsset{compat=newest}
\DeclarePairedDelimiterX{\norm}[1]{\lVert}{\rVert}{\ifblank{#1}{\:\cdot\:}{#1}}
\DeclarePairedDelimiterX{\abs}[1]{\lvert}{\rvert}{\ifblank{#1}{\:\cdot\:}{#1}}

\newcommand{\SetSymbol}[1][]%
{\nonscript\:#1\vert\allowbreak\nonscript\:\mathopen{}}
\DeclarePairedDelimiterX{\set}[1]{\lbrace}{\rbrace}{%
  #1}
\DeclareMathOperator*{\argmin}{arg\,min}
\title{%
  Clustering-Based Quantisation for\\%
  PDE-Based Image Compression%
}
\titlerunning{%
  Efficient Quantisation for PDE-Based Image Compression
}
\author{%
  Laurent Hoeltgen \and%
  Pascal Peter     \and%
  Michael Breuß%
}
\authorrunning{%
  Hoeltgen et al.%
}
\institute{
  L. Hoeltgen \and M. Breuß
  \at{} Chair for Applied Mathematics, BTU Cottbus-Senftenberg, Cottbus, Germany,
  \email{$\lbrace$hoeltgen,breuss$\rbrace$@b-tu.de}
  \and
  P. Peter%
  \at{} Mathematical Image Analysis Group, Saarland University, Saarbrücken,
  Germany, \email{peter@mia.uni-saarland.de}
}
\journalname{\mbox{}}
\date{\mbox{}}
\begin{document}
\maketitle
\begin{abstract}
      Finding optimal data for inpainting is a key problem in the context of
      partial differential equation based image compression. The data that
      yields the most accurate reconstruction is real-valued. Thus, quantisation
      models are mandatory to allow an efficient encoding. These can also be
      understood as challenging data clustering problems. Although clustering
      approaches are well suited for this kind of compression codecs, very few
      works actually consider them. Each pixel has a global impact on the
      reconstruction and optimal data locations are strongly correlated with
      their corresponding colour values. These facts make it hard to predict
      which feature works best.\par
      In this paper we discuss quantisation strategies based on popular methods
      such as k-means. We are lead to the central question which kind of feature
      vectors are best suited for image compression. To this end we consider
      choices such as the pixel values, the histogram or the colour map.\par
      Our findings show that the number of colours can be reduced significantly
      without impacting the reconstruction quality. Surprisingly, these benefits
      do not directly translate to a good image compression performance. The
      gains in the compression ratio are lost due to increased storage costs.
      This suggests that it is integral to evaluate the clustering on both, the
      reconstruction error and the final file size.
\end{abstract}
%
\section{Introduction}
\label{sec:introduction}
%
A major challenge in data analysis is the reconstruction of a signal from very
few data points. In image processing this interpolation problem is called
inpainting. Recent advances have shown that accurate reconstructions from a
small sample of well chosen image points are possible by using methods based on
partial differential equations
(PDEs)~\cite{GWWB05,LSWL07,BBBW2009,SWB2009,MHWT2012,BU13,HSW2013,CRP2014}.
These efforts have also been exploited for lossy image compression schemes that
can nowadays compete with other state-of-the-art approaches
\cite{PSMM15,SWB2009}. Even simple methods that rely on the interpolation
capabilities of the Laplacian may yield more accurate reconstructions than
JPEG~2000~\cite{PHHN2015}. Unfortunately, the underlying optimisation problem
concerned with selecting optimal point locations is computationally intensive.
In addition, the resulting data is often difficult to store efficiently. The
pixel locations may be scattered arbitrarily inside the image domain and the
corresponding colour values are often computed in floating point precision.
Storing this data as-is is prohibitively expensive and often unnecessary. The
human visual system is only capable of distinguishing about thirty different
shades of grey~\cite{KMHD2013}. These thirty colours can be encoded with 5~bit
instead of the 64~bit required for a floating point value in double precision.
Finding the optimal number and distribution of these grey values is however a
complicated task. Nevertheless, it is reasonable to expect that small changes to
the colour values will only yield small changes in the reconstruction. This
assumption allows us to replace real valued colours by their closest integer
valued neighbours without causing to much damage.\par
This work extends the preliminary findings from~\cite{HB2016}. There, the
authors investigated clustering schemes and quality assessment measures to
provide fast strategies to determine a good number of quantised colours. In
\cite{HB2016}, the clustering was not evaluated against the compression ratio of
the final compressed file but only against the reconstruction quality of the
clustered data. Nevertheless, the results indicated that simple clustering
approaches yield good quantisations of the co-domain. In addition, the authors
showed that these clusterings can be evaluated with quality measures such as the
silhouette coefficient.\par
%
\textbf{Our Contribution.}
%
In this paper we investigate how well different clustering optimisation
techniques fare in a data compression context. Thus, we extend the results from
\cite{HB2016} by providing a similar evaluation on the compression ratio. In
addition we rank all considered clustering strategies with respect to their
effectiveness and show that the quantisation must be done with the compression
ratio in mind and not with respect to the reconstruction quality.\par
To this end we will discuss several algorithms as well as quality measures and
show that not only the number of clusters, but also the final distribution of
the labels is important. The distribution of the labels is mostly influenced by
the underlying clustering algorithm, whereas the number of clusters can be
optimised independently for any strategy. The latter task is also considered in
detail. Several strategies are proposed. This discussion also extends the
findings from~\cite{HB2016} where only the silhouette coefficient was taken into
consideration.\par
%
\section{Partial Differential Equation Based Image Compression}
\label{sec:compression}
%
Even though most diffusion-type partial differential equations can be used to
reconstruct the data, previous works have shown that homogeneous, linear
diffusion ranks among the best performing ones~\cite{PHHN2015} in the
compression context. We also refer to \cite{B2015} for a more general overview
on PDE-based inpainting methods. The simplicity of homogeneous diffusion
inpainting allows a thorough mathematical analysis and an extensive
optimisation~\cite{HSW2013,HW15}. Since our sought optimal quantisation is
likely to depend strongly on the underlying reconstruction process we focus
exclusively on homogeneous diffusion in this work. Nevertheless, many of the
presented strategies can be extended to other inpainting methods.\par
A good image compression approach consists of several components. Besides the
reconstruction method it is also necessary to optimise the considered data and
to design an efficient encoding. Let us now briefly review the involved
reconstruction and optimisation steps in the next paragraphs.
%
\subsection{Inpainting with Homogeneous Diffusion}
\label{cha:inpa-with-homog}
%
Inpainting with homogeneous diffusion (sometimes also called Laplace
interpolation) is a rather simple reconstruction method that is well suited for
highly scattered data in arbitrary dimensional settings. It can be modelled as
follows. Let $f:\Omega\to\mathbb{R}$ be a smooth function on some bounded domain
$\Omega\subset\mathbb{R}^2$ with a sufficiently regular boundary
$\partial\Omega$. Throughout this work, we will restrict ourselves to the case
of grey value images even though many of the results hold for arbitrary numbers
of colour channels. Moreover, let us assume that there exists a closed nonempty
set of known data $\Omega_K\subsetneqq\Omega$ that will be interpolated by the
underlying diffusion process. Homogeneous diffusion inpainting considers the
following partial differential equation with mixed boundary conditions:
\begin{equation}
  \label{eq:mixed-boundary-problem}
  \begin{split}
        -&\Delta u = 0\quad\quad\ \text{on}\ \Omega\setminus\Omega_K\\
        \text{with}\quad
        &\begin{cases}
              \phantom{\partial_n}u=f\ &\text{on}\
              \partial\Omega_K\\
              \partial_n u = 0\ &\text{on}\
              \partial\Omega\setminus{}\partial\Omega_K
        \end{cases}
  \end{split}
\end{equation}
where $\partial_n u$ denotes the derivative of $u$ in the outer normal
direction. We assume that both boundary sets $\partial\Omega_K$ and
$\partial\Omega \setminus{} \partial\Omega_K$ are non-empty. Equations of this
type are commonly referred to as mixed boundary value problems and sometimes
also as Zaremba's problem \cite{Z1910}. The existence and uniqueness of
solutions has been studied during the last century. Showing that
\eqref{eq:mixed-boundary-problem} is indeed solvable is by no means a trivial
feat. We refer to~\cite{GT2001} for an extensive study of linear elliptic
partial differential equations and to~\cite{HHBK2017} for a more particular
analysis of \eqref{eq:mixed-boundary-problem}. A particularly simple case is the
1-D setting, where the solution can obviously be expressed using piecewise
linear splines interpolating data on $\partial\Omega_K$.\par{}
%
\subsection{Optimal Inpainting Positions}
\label{sec:optknt}
%
Finding good inpainting positions is a task related to the free knot problem
from interpolation theory~\cite{B1974a,H2002}. In the special context of image
inpainting and image compression the authors of
\cite{CRP2014,MHWT2012,HSW2013,DI2004} suggested several strategies to find good
locations. Mainberger et al.~\cite{MHWT2012} propose stochastic optimisations
that bear similarities to simulated annealing approaches. Chen et
al.~\cite{CRP2014} use fast bi-level optimisation schemes and in~\cite{HSW2013}
Hoeltgen et al.\ present an optimal control model. Finally the works of Demaret
et al.~\cite{DI2003,DI2004,DIK2009} use adaptive triangulations and mesh
optimisation strategies, whereas Ochs et al.~\cite{OCBP2014} suggest fast
numerics for the task at hand.
%
\subsection{Continuous Grey Value Optimisation}
\label{sec:gvo}
%
In~\cite{MHWT2012,HW15} the authors discuss approaches to find the best pixel
values. These algorithms have in common, that they all yield real valued
floating point results. If we denote the reconstruction operator that solves
\eqref{eq:mixed-boundary-problem} for a given mask $\Omega_{K}$ by
$M\left( \Omega_{K} \right)$, then this problem can be formulated as
\begin{equation}
  \label{eq:gvo-equation}
  \argmin_{u}\,\set*{\norm*{M\left( \Omega_{K} \right) u -f}^{2}}
\end{equation}
Since our initial PDE is linear, the task stated in~\eqref{eq:gvo-equation}
corresponds to a linear least squares problem and can be solved efficiently. An
alternative discrete optimisation has been suggested by Schmaltz et
al.~\cite{SWB2009}. The latter approach iterates over the inpainting data and
selects in a greedy way the currently best quantised grey value. The iterations
continue until a convergent state is reached. While this approach comes
conceptually close to clustering strategies, it has two notable drawbacks:
\emph{(i)} The approach requires that the distribution of the quantised values
is fixed at the beginning, i.e.\ it does not change during the iterations;
\emph{(ii)} to test a certain grey value for its quality requires the solution
of a large and sparse linear system, i.e.\ the computational costs are
considerable as the algorithm requires the solution of thousands of linear
systems. Nevertheless, we remark that it yields excellent qualitative
results.\par
%
\section{Clustering Inpainting Data}
\label{sec:cluster}
%
The following paragraphs discuss potential choices for the feature selection as
well as for the algorithmic execution. Our motivation is to use models that come
conceptually close to the continuous grey value optimisation and the
quantisation aware approach of Schmaltz et al.~\cite{SWB2009}. Therefore, we
focus on the squared Euclidean distance as cost function and seek the best set
of grey values for the mask pixel locations. However, we will refrain from using
the reconstruction inside our algorithms in order to obtain fast methods.
%
\subsection{Feature Selection}
\label{sec:feature}
%
Our setup provides us the full image data $f$ as well as an optimised inpainting
mask $\Omega_{K}$. There exist many possible choices to extract interesting
features from this data. The following eight feature choices seem reasonable:
\begin{enumerate}
\item position and value of all image pixels
\item position and value of all mask pixels
\item grey values values of all image pixels (i.e.\ discarding the positional
      information)
\item grey values of all mask pixels (i.e.\ discarding the positional
      information)
\item the histogram of all image pixels
\item the histogram of all mask pixels
\item the colour map (i.e.\ histogram without frequency count) of all image
      pixels
\item the colour map (i.e.\ histogram without frequency count) of all mask
      pixels
\end{enumerate}
The following observations can be made with respect to the feature choices.
\begin{itemize}
\item We remark that including the pixel position into the feature vector
      renders each sample unique. In that case it makes no sense to track their
      frequency as it will always be 1.
\item Approaches that exploit the full image information have better chances at
      adapting to features in the image that are not covered by the mask pixels
      alone. On the other hand, optimised positions for the inpainting with
      homogeneous diffusion are usually in the vicinity of edges and other
      important structures. Therefore, it is likely that the image information
      provided by the mask pixels is already sufficient to determine a good
      clustering. In addition, for compression purposes the inpainting masks are
      usually very sparse. Thus, we also benefit from a smaller feature set and
      a higher performance.
\item Strategies that focus exclusively on the grey values of an image are
      likely to perform fastest since they can be carried out on the histogram.
      As soon as we include the spatial position into the feature set we obtain
      as many unique feature values as pixels. For large images we may encounter
      memory or run time restrictions.
\end{itemize}
Finally let us remark that from a clustering perspective it is irrelevant
whether we consider grey values or colour images. Only the mask optimisation and
inpainting steps are more difficult to carry out for colour images.
%
\subsection{Clustering Approaches for Image Quantisation}
\label{sec:clusteralgs}
%
In this work we opt for some of the most popular and best understood approaches
to classify our data. Since clustering approaches are commonly subdivided into
partitional and hierarchical methods we select the following representatives:
\begin{enumerate}
\item \emph{partitional clustering:} We use the k-means++ variant of the
      venerable k-means algorithm of Lloyd~\cite{L1982}.
\item \emph{hierarchical clustering:} We employ a bottom up strategy where
      initially each feature represents a single cluster. These clusters are
      merged step-by-step using Ward's criterion~\cite{W1963}.
\item \emph{probabilistic clustering:} As an alternative to the ``hard''
      labelling performed by the k-means approach we also consider a ``softer''
      variant using a Gaussian mixture model and employ the popular expectation
      maximisation algorithm of Sundberg~\cite{S1974,S1976} to compute the
      probabilities that a feature belongs to a certain cluster.
\end{enumerate}
We also refer to~\cite{GRBL2012} for a discussion on the viability of various
clustering techniques.\par
%
\subsection{Quality Measures and Optimal Number of Clusters}
\label{sec:quality}
%
When evaluating a partitioning of the data we need to consider several criteria.
First, the quality of the clustering itself: Features in the same group should
indeed be similar and yet also distinct from observations from other clusters.
For our purposes it is also important to assess the quality of the
reconstruction as well as the compression ratio of the final file. Especially
the latter quantity is difficult to estimate. Our quantised data should have few
different grey values, (i.e.\ a small number of clusters) such that the file
compression can be efficient. However, it is also essential that the
reconstruction quality is fair, which is likely to require many different grey
values (i.e.~a large number of clusters) in floating point precision. Thus,
there is a trade-off between reconstruction quality and compression ratio. This
influence can be steered by optimising the number of clusters. From the large
pool of quality measures available in the literature (\cite{D2013} lists more
than 50) we chose the following to help us identify an optimal grouping of our
data. In this description, $k$ denotes always the number of clusters and $N$ the
number of features.
\begin{enumerate}
\item The Calinsky Harabasz (CH) criterion~\cite{CH1974}, also called
      \emph{variance ratio criterion}, considers the quantity
      \begin{equation}
            \label{eq:2}
            \frac{\sum_{j=1}^{k} \norm*{m_{j}-m}^{2}}
            {\sum_{j=1}^{k} \sum_{x\in C_{j}} \norm{x-m_{j}}^{2}}
            \frac{N-k}{k-1}
      \end{equation}
      where $m_{j}$ is the centroid of the cluster $C_{j}$ and $m$ the overall
      mean of all features. Higher values indicate better clusterings.
\item The Davies Bouldin (DB) criterion~\cite{DB1979} is based on a ratio of
      within-cluster and between-cluster distances. The Davies-Bouldin index is
      defined as
      \begin{equation}
            \label{eq:4}
            \frac{1}{k} \sum_{j=1}^{k} \max_{i\neq j}
            \set*{\frac{\bar{d}_{j}+\bar{d}_{i}}{d_{j,i}}}
      \end{equation}
      with $\bar{d_{\ell}}$ being the average distance of the centroid to each
      element in the cluster $C_{\ell}$ and where $d_{r,s}$ is the distance
      between the centroids of the clusters $C_{r}$ and $C_{s}$. Lower values
      are better.
\item The GAP criterion~\cite{TWH2001} formalises the well known \emph{L-term
        heuristic} by estimating the ``elbow'' location. The ``elbow'' occurs at
      the most dramatic decrease in error measurement, i.e.\ the largest gap
      value. To this end, the within cluster sum of squares is compared to its
      expectation under an appropriate null reference distribution of the data.
      This leads to the following formula for the GAP value:
      \begin{equation}
            \label{eq:1}
            \begin{aligned}
                  GAP_{N}(k) &\coloneqq\mathrm{E}_{N}^{*}[\log (W_{k} )] -
                  \log (W_{k}) \\
                  W_{k} &\coloneqq \sum_{r=1}^{k} \frac{1}{2\abs*{C_{r}}}
                  \sum_{c_{i},c_{j}\in C_{r}} \norm*{c_{i}-c_{r}}
            \end{aligned}
      \end{equation}
      where $E^{*}_{N}$ denotes the expectation under a sample of size $N$ from
      the reference distribution. The optimal number of clusters is given by $k$
      which maximises the GAP value.
\item The Silhouette index~\cite{R1987} of a cluster $C$ is defined to be the
      average silhouette value of all its elements. The silhouette value of a
      single element $x\in C$ is defined as
      \begin{equation}
            \label{eq:3}
            \begin{cases}
                  0, &\text{if}\ \mathrm{dist}(x,C) = 0\\
                  \frac{\mathrm{dist}(x,B)-\mathrm{dist}(x,C)}
                  {\max\set*{\mathrm{dist}(x,C),\mathrm{dist}(x,B)}},
                  &\text{else}
            \end{cases}
      \end{equation}
      where $\mathrm{dist}(x,C)$ is the distance of $x$ to the cluster $C$ and
      where $B$ is the next closest cluster. Comparing the silhouette indices
      for all possible clusters gives an estimate on how well structured the
      clustering is. These indices should be as close to 1 as possible.
\end{enumerate}
We remark that the GAP criterion is one of the most flexible approaches but also
the most expensive measure in terms of computational effort.
%
\section{Numerical Experiments}
\label{sec:experiments}
%
\subsection{Performance of the Clustering Algorithms}
\label{sec:PerClustAlgs}
%
In total the proposed 8 features, 3 algorithms and 6 quality criteria allow 144
possible combinations to be evaluated. An exhaustive testing would require the
evaluation of all these approaches on several test images, thus increasing the
data to be analysed even more. Certainly not all of these combinations are
reasonable and some strategies also require considerable run times such that
they become impractical for most applications. Thus, we have decided to restrict
ourselves to the following choices. This list is also motivated by the findings
from~\cite{HB2016} where it has been noticed that the k-means and hierarchical
clustering are more reliable and usually perform better than the probabilistic
clustering with a Gaussian mixture model.
\begin{enumerate}
\item k-means on the position and value of all mask pixels\label{item:1}
\item hierarchical clustering on the position and value of all mask
      pixels\label{item:2}
\item probabilistic clustering on the position and value of all mask
      pixels\label{item:3}
\item k-means on the pixel values of all mask pixels\label{item:4}
\item hierarchical clustering on the pixel values of all mask
      pixels\label{item:5}
\item k-means on the histogram of all mask pixels\label{item:7}
\item k-means on the histogram of all image pixels\label{item:8}
\item k-means on the unique grey values of all mask pixels\label{item:9}
\item k-means on the unique grey values of all image pixels\label{item:10}
\end{enumerate}
Our experiments were carried out on the data set presented in
Fig.~\ref{fig:trui}. The considered test image is of size $256\times{}256$ and
has 170 unique grey values. These are byte-wise coded (i.e.~the maximal amount
of different grey values is 256) and the corresponding binary mask has a density
of~5\%. It has been obtained with the optimal control approach of Hoeltgen et
al.~\cite{HSW2013}. The histograms of the image and of the pixels indicated by
the mask are depicted in Fig.~\ref{fig:truiHist}. The considered test image and
its histogram are representative for a large class of images. The image has
smooth regions as well as highly textured areas. The corresponding histograms
are also very generic. They do not show any particular patterns. As a such, we
believe that the presented findings would be similar for other images. The
methods presented in this work are also agnostic towards how the mask was
obtained.\par
\begin{figure}
      \centering
      \includegraphics[width=0.2\textwidth]{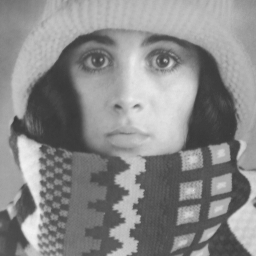}\hspace*{5mm}
      \includegraphics[width=0.2\textwidth]{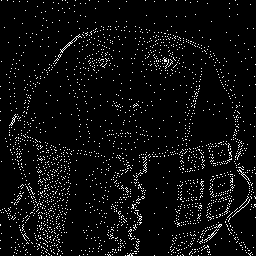}
      \caption{\emph{Left}: The considered test image trui. The image is
        $256\times{}256$ in size. \emph{Right}: A corresponding inpainting mask
        with a density of $5\%$. Locations used for the reconstruction are
        marked in white.}
      \label{fig:trui}
\end{figure}
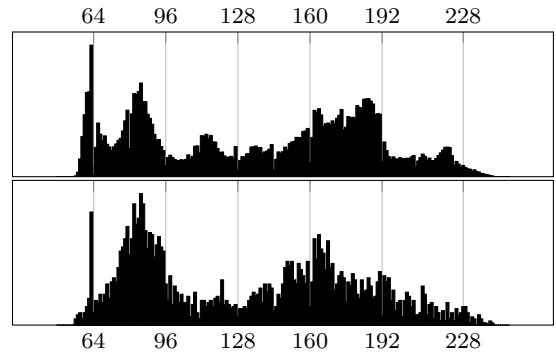
\begin{figure}
      \centering
      \begin{tikzpicture}
            \begin{axis}[%
                  /tikz/ybar,
                  xtick align=inside,
                  ymin=0,
                  bar width=1.1pt,
                  grid=major,
                  height=3.5cm,
                  xtick={64, 96, 128, 160, 192, 228},
                  xticklabel pos=upper,
                  ytick=\empty,
                  width=0.5\textwidth]
                  \addplot[black,fill=black] coordinates {(48, 0) (49, 0) (50, 0)
                    (51, 0) (52, 0) (53, 0) (54, 0) (55, 0) (56, 8) (57, 47)
                    (58, 183) (59, 414) (60, 640) (61, 871) (62, 878) (63, 1361)
                    (64, 16) (65, 304) (66, 555) (67, 439) (68, 373) (69, 422)
                    (70, 335) (71, 307) (72, 271) (73, 309) (74, 338) (75, 376)
                    (76, 397) (77, 473) (78, 577) (79, 688) (80, 287) (81, 650)
                    (82, 859) (83, 850) (84, 882) (85, 969) (86, 776) (87, 775)
                    (88, 646) (89, 603) (90, 542) (91, 438) (92, 376) (93, 384)
                    (94, 311) (95, 269) (96, 112) (97, 198) (98, 222) (99, 197)
                    (100, 178) (101, 168) (102, 163) (103, 176) (104, 164) (105,
                    179) (106, 226) (107, 182) (108, 211) (109, 314) (110, 318)
                    (111, 243) (112, 424) (113, 412) (114, 439) (115, 390) (116,
                    427) (117, 357) (118, 358) (119, 291) (120, 243) (121, 255)
                    (122, 215) (123, 202) (124, 193) (125, 210) (126, 191) (127,
                    312) (128, 60) (129, 166) (130, 207) (131, 201) (132, 236)
                    (133, 234) (134, 306) (135, 307) (136, 317) (137, 295) (138,
                    301) (139, 241) (140, 256) (141, 265) (142, 284) (143, 332)
                    (144, 71) (145, 179) (146, 256) (147, 241) (148, 258) (149,
                    306) (150, 273) (151, 343) (152, 359) (153, 365) (154, 407)
                    (155, 404) (156, 400) (157, 467) (158, 473) (159, 499) (160,
                    140) (161, 406) (162, 686) (163, 657) (164, 703) (165, 639)
                    (166, 520) (167, 547) (168, 555) (169, 536) (170, 577) (171,
                    543) (172, 604) (173, 635) (174, 696) (175, 486) (176, 512)
                    (177, 621) (178, 606) (179, 615) (180, 600) (181, 735) (182,
                    713) (183, 709) (184, 801) (185, 803) (186, 807) (187, 781)
                    (188, 752) (189, 642) (190, 651) (191, 647) (192, 157) (193,
                    359) (194, 271) (195, 206) (196, 171) (197, 205) (198, 185)
                    (199, 174) (200, 196) (201, 183) (202, 186) (203, 191) (204,
                    186) (205, 203) (206, 219) (207, 97) (208, 162) (209, 197)
                    (210, 234) (211, 211) (212, 163) (213, 192) (214, 166) (215,
                    161) (216, 188) (217, 218) (218, 246) (219, 274) (220, 303)
                    (221, 306) (222, 301) (223, 217) (224, 116) (225, 158) (226,
                    129) (227, 95) (228, 116) (229, 94) (230, 79) (231, 77)
                    (232, 56) (233, 71) (234, 48) (235, 48) (236, 34) (237, 29)
                    (238, 19) (239, 19) (240, 10) (241, 7) (242, 0) (243, 0)
                    (244, 0) (245, 0) (246, 0) (247, 0) (248, 0)};
            \end{axis}
      \end{tikzpicture}
      \begin{tikzpicture}
            \begin{axis}[%
                  /tikz/ybar,
                  xtick align=inside,
                  ymin=0,
                  bar width=1.1pt,
                  grid=major,
                  height=3.5cm,
                  xtick={64, 96, 128, 160, 192, 228},
                  ytick=\empty,
                  width=0.5\textwidth]
                  \addplot[black,fill=black] coordinates {(48, 0) (49, 0) (50, 0)
                    (51, 0) (52, 0) (53, 0) (54, 0) (55, 0) (56, 3) (57, 5) (58,
                    6) (59, 3) (60, 6) (61, 9) (62, 13) (63, 55) (64, 6) (65,
                    12) (66, 11) (67, 15) (68, 11) (69, 15) (70, 20) (71, 13)
                    (72, 21) (73, 23) (74, 21) (75, 28) (76, 36) (77, 38) (78,
                    42) (79, 48) (80, 34) (81, 42) (82, 59) (83, 49) (84, 52)
                    (85, 64) (86, 59) (87, 46) (88, 37) (89, 45) (90, 44) (91,
                    33) (92, 39) (93, 43) (94, 38) (95, 36) (96, 20) (97, 12)
                    (98, 24) (99, 10) (100, 19) (101, 16) (102, 11) (103, 15)
                    (104, 11) (105, 11) (106, 14) (107, 4) (108, 10) (109, 16)
                    (110, 6) (111, 3) (112, 12) (113, 12) (114, 9) (115, 11)
                    (116, 12) (117, 9) (118, 13) (119, 14) (120, 11) (121, 22)
                    (122, 10) (123, 12) (124, 10) (125, 8) (126, 9) (127, 10)
                    (128, 4) (129, 8) (130, 8) (131, 9) (132, 8) (133, 12) (134,
                    14) (135, 16) (136, 14) (137, 15) (138, 18) (139, 11) (140,
                    20) (141, 16) (142, 20) (143, 18) (144, 7) (145, 9) (146,
                    15) (147, 13) (148, 20) (149, 28) (150, 31) (151, 29) (152,
                    31) (153, 22) (154, 20) (155, 30) (156, 27) (157, 24) (158,
                    20) (159, 31) (160, 9) (161, 21) (162, 42) (163, 32) (164,
                    44) (165, 37) (166, 35) (167, 27) (168, 41) (169, 28) (170,
                    30) (171, 17) (172, 20) (173, 24) (174, 26) (175, 23) (176,
                    11) (177, 21) (178, 19) (179, 25) (180, 22) (181, 15) (182,
                    13) (183, 22) (184, 16) (185, 17) (186, 23) (187, 19) (188,
                    13) (189, 14) (190, 21) (191, 26) (192, 4) (193, 24) (194,
                    21) (195, 23) (196, 12) (197, 17) (198, 7) (199, 11) (200,
                    12) (201, 4) (202, 12) (203, 16) (204, 14) (205, 9) (206,
                    12) (207, 5) (208, 5) (209, 6) (210, 19) (211, 15) (212, 5)
                    (213, 10) (214, 11) (215, 3) (216, 9) (217, 6) (218, 2)
                    (219, 7) (220, 2) (221, 11) (222, 3) (223, 6) (224, 2) (225,
                    8) (226, 5) (227, 6) (228, 9) (229, 9) (230, 6) (231, 1)
                    (232, 4) (233, 3) (234, 0) (235, 5) (236, 3) (237, 1) (238,
                    1) (239, 0) (240, 2) (241, 1) (242, 0) (243, 0) (244, 0)
                    (245, 0) (246, 0) (247, 0) (248, 0)};
            \end{axis}
      \end{tikzpicture}
      \caption{\emph{Top:} Histogram of the trui test image. \emph{Bottom:}
        Histogram of the data at the mask positions.}
      \label{fig:truiHist}
\end{figure}
Let us in a first step evaluate our considered criteria on the trui test image
from Fig.~\ref{fig:trui}. The results of our experiments are listed in
Table~\ref{tab:ClustResults}. We see that integrating the position into the
feature vectors significantly deteriorates the reconstruction quality (rows 1-3
in Table~\ref{tab:ClustResults}). The suggested optimal numbers of clusters
however look reasonable, except for the GAP criterion which tends to suggest
rather large numbers of clusters. The other experiments (rows 4-9 in
Table~\ref{tab:ClustResults}) yield, at least in terms of mean squared error,
good results. Also here the optimal number of clusters is often overestimated.
Notable exceptions are the GAP criterion for the setups~\ref{item:4},
\ref{item:5}, \ref{item:9} and~\ref{item:10} where the suggested number of
clusters was around 12 and the reconstruction error around 52. For the setup
\ref{item:7}, the GAP criterion suggested 36 colours with a corresponding mean
squared error of 48.91. Considering that the reconstruction with the original
170 colours has an mean squared error of 46.96, this is actually a very
satisfying result. Also the suggested number of clusters coincides with our
expectations.\par
\begin{figure}
      \centering
      \begin{tikzpicture}
            \begin{axis}[ymin=46, ymax=58, xmin=10, xmax=74,
                  width=8.5cm,
                  height=4cm,
                  xtick = {12,20,30,40,50,60,72},
                  ytick = {46,48,50,52,54,56,58},
                  xlabel={number of clusters},
                  ylabel={mean squared error},
                  grid = major,
                  legend entries={k-means, hierarchical, original data},
                  legend cell align=left,
                  legend style={legend pos=north east},
                  ]
                  \addplot+[mark=none, dotted,   very thick, red]
                  table[x index=0, y index=1] {data/mse-12-4-1.dat};
                  \addplot+[mark=none, dashed, very thick, blue]
                  table[x index=0, y index=2] {data/mse-12-4-1.dat};
                  \addplot [mark=none, solid, color=black, very thick]
                  coordinates {(10,46.96) (75,46.96)};
            \end{axis}
      \end{tikzpicture}
      \caption{Evolution of the mean squared error of the reconstruction from
        the clustered pixel values on the mask position. The solid black line
        marks the reconstruction error for the unaltered data. As we can see,
        the k-means method performs slightly better. For 35 or more clusters
        there is hardly any improvement in the reconstruction anymore. Also the
        mean squared error converges towards the error obtained by using the
        original data. This suggests, that the optimal number of clusters should
        actually be somewhere between 30 and 40 clusters.}
      \label{fig:MSEClust}
\end{figure}
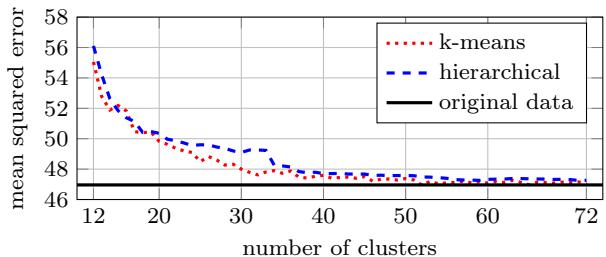

\begin{table*}
      \centering
      \caption{%
        Clustering results for the trui test image. We tested all cluster sizes
        ranging from 12 to 72. The mean squared error is always given for the
        corresponding suggested optimal number of clusters. As we can see, most
        methods tend to return cluster numbers close to the maximal or minimal
        allowed value. Features that include the position of a pixel yield
        rather bad results. The suggested optimal number of clusters from the
        GAP criterion coincides best with our expectations. The corresponding
        reconstruction errors are also quite good, considering that the
        reconstruction error with the original 170 colours had an error of
        46.96}\label{tab:ClustResults}
      \begin{tabular*}{\textwidth}{@{\extracolsep{\fill}}@{}ccccccccc@{}}
        \toprule \addlinespace
        \multirow{2}{*}{Feature} &
        \multicolumn{4}{c}{mean squared error} &
        \multicolumn{4}{c}{Best number of clusters} \\
        \cmidrule{2-5} \cmidrule{6-9}
        & Silhouette & DB & CH & GAP
        & Silhouette & DB & CH & GAP \\
        \addlinespace
        \cmidrule{1-1}
        \cmidrule{2-2} \cmidrule{3-3} \cmidrule{4-4} \cmidrule{5-5}
        \cmidrule{6-6} \cmidrule{7-7} \cmidrule{8-8} \cmidrule{9-9}
        \addlinespace
        \ref{item:1} & 404 & 246 & 350 & 125 & 16 & 25 & 15 & 66 \\
        \ref{item:2} & 380 & 199 & 271 & 147 & 13 & 38 & 27 & 68 \\
        \ref{item:3} & 418 & 316 & 295 & 152 & 12 & 15 & 19 & 71 \\
        \addlinespace
        \ref{item:4} & 47.15 & 47.08 & 46.95 & 54.88 & 63 & 72 & 70 & 12 \\
        \ref{item:5} & 47.26 & 47.32 & 47.26 & 52.63 & 72 & 69 & 72 & 14 \\
        \addlinespace
        \ref{item:7} & 63.17 & 47.51 & 47.66 & 48.91 & 12 & 72 & 72 & 36 \\
        \ref{item:8} & 52.50 & 51.65 & 53.01 & 54.86 & 72 & 72 & 72 & 66  \\
        \addlinespace
        \ref{item:9}  & 51.30 & 46.63 & 46.86 & 50.34 & 12 & 63 & 69 & 12 \\
        \ref{item:10} & 52.94 & 46.57 & 46.62 & 49.88 & 12 & 72 & 67 & 12 \\
        \addlinespace\bottomrule
      \end{tabular*}
\end{table*}
Figure~\ref{fig:MSEClust} plots the evolution of the reconstruction error as a
function of the number of clusters for the k-means method as well as for the
hierarchical clustering. The k-means method performs slightly better than the
hierarchical scheme. For 35 or more clusters there is hardly any improvement in
the reconstruction anymore and the errors converge towards the mean squared
error of the original data. The steady and rather stable decrease of the error
might explain why most clustering methods tend to overestimate the number of
clusters. In terms of error, the findings for 40 clusters are almost as good as
those with 72 clusters. Without an explicit requirement that the optimal number
of clusters should be small, these results are difficult to discern.
%
\subsection{Application to Image Compression}
\label{sec:appl-image-compr}
%
In order to evaluate the impact of quantisation by clustering on actual
compression applications, we combine our approach with the state-of-the-art
codec of Peter et al.~\cite{PHHW16}. It stores binary masks with free choice of
locations on the pixel grid by employing a block coding scheme~\cite{ZA89}. In
the original codec, the corresponding grey values are subject to an equidistant
quantisation and both the locations and values are finally stored with the
context-mixing scheme PAQ by Mahoney~\cite{Ma05}. In the following, we
investigate how replacing the equidistant quantisation with our clustering
approach influences the performance of this codec. Since kmeans++ has yielded
the most consistent results in the previous sections, we apply this strategy in
the following.\par
\begin{figure*}
      \centering
      \begin{center}
            \begin{tikzpicture}
                  \begin{axis}[
                        width=11cm,
                        height=5cm,
                        ymin=14, ymax=60,
                        xmin=14, xmax=19,
                        xtick = {15, 16, 17, 18},
                        ytick = {20, 25, 30, 35, 40, 45, 50, 55},
                        xlabel={compression ratio},
                        ylabel={mean squared error},
                        grid = major,
                        legend entries={
                          equidistant,
                          k-means,
                          equidistant with tonal opt.,
                          k-means with tonal opt.},
                        legend cell align=left,
                        legend style={legend pos= outer north east},
                        ]
                        \addplot+[mark=none, solid, very thick, red]
                        table[x index=4, y index=5] {data/trui-1-3.tab};
                        \addplot+[mark=none, dotted, very thick, blue]
                        table[x index=0, y index=1] {data/trui-1-3-sorted.csv};
                        \addplot+[mark=none, dashed, very thick, red]
                        table[x index=4, y index=3] {data/truigvo-1-3.tab};
                        \addplot+[mark=none, dashdotted, very thick, blue]
                        table[x index=0, y index=1] {data/truigvo-1-3-sorted.csv};
                  \end{axis}
            \end{tikzpicture}
      \end{center}
      \caption{Comparison of compression performance on \emph{trui} over
        different compression ratios. The solid and dotted curves show the
        results with quantised original grey values whereas the dashed and
        dash-dotted curves present the findings for the compression with
        quantisation-aware tonal optimisation.}
      \label{fig:Perf}
\end{figure*}
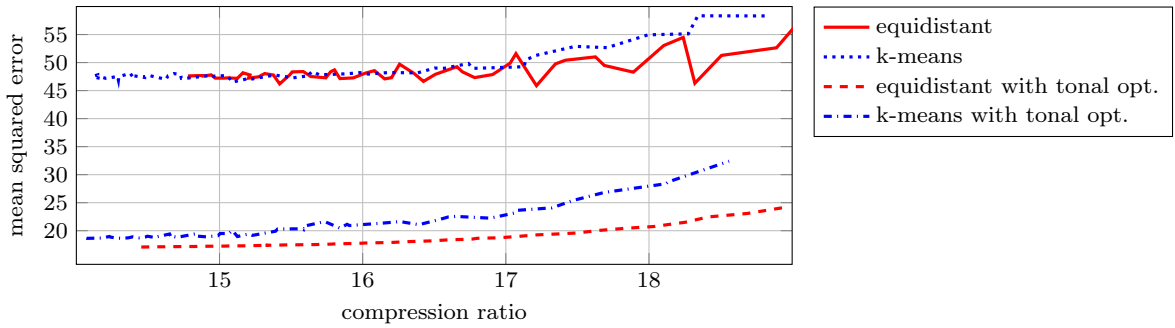
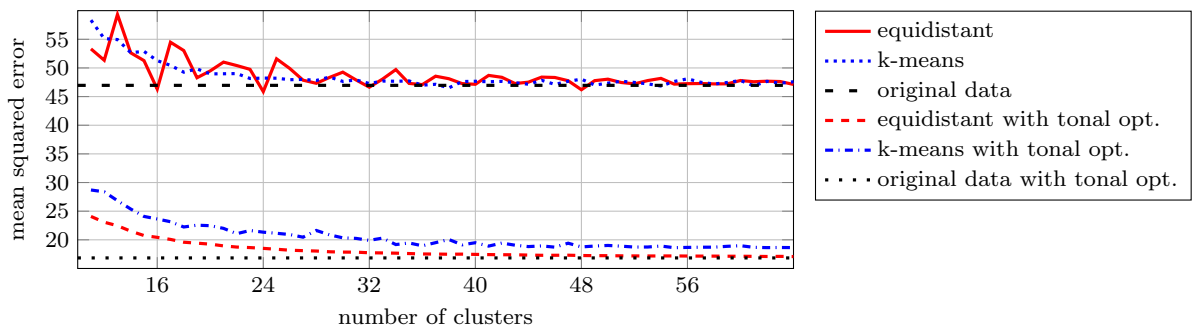
\begin{figure*}
      \centering
      \begin{tikzpicture}
            \begin{axis}[
                  width=11cm,
                  height=5cm,
                  xmin=10, xmax=64,
                  ymin=15, ymax=60,
                  xtick = {16,24,32,40,48,56},
                  ytick = {20, 25, 30, 35, 40, 45, 50, 55},
                  xlabel={number of clusters},
                  ylabel={mean squared error},
                  grid = major,
                  legend entries={
                    equidistant,
                    k-means,
                    original data,
                    equidistant with tonal opt.,
                    k-means with tonal opt.,
                    original data with tonal opt.},
                  legend cell align=left,
                  legend style={legend pos=outer north east},
                  ]
                  \addplot+[mark=none, solid, very thick, red]
                  table[x index=6, y index=5] {data/trui-1-3.tab};
                  \addplot+[mark=none, dotted, very thick, blue]
                  table[x index=6, y index=1] {data/trui-1-3.tab};
                  \addplot [mark=none, loosely dashed, color=black, very thick]
                  coordinates {(10,46.96) (75,46.96)};
                  \addplot+[mark=none, dashed, very thick, red]
                  table[x index=6, y index=3] {data/truigvo-1-3.tab};
                  \addplot+[mark=none, dashdotted, very thick, blue]
                  table[x index=6, y index=1] {data/truigvo-1-3.tab};
                  \addplot [mark=none, loosely dotted, color=black, very thick]
                  coordinates {(10,16.84) (75,16.84)};
            \end{axis}
      \end{tikzpicture}
      \caption{Performance evaluation of our quantisation schemes in terms of
        reconstruction error. The solid and dotted curves denote the findings
        for an equidistant quantisation and a k-means quantisation without a
        tonal optimisation as post processing. The other two curves depict the
        findings with an additional tonal optimisation. The black lines denote
        the error when using the original data and tonal optimised data.}
      \label{fig:PlotInpaintErr}
\end{figure*}
\begin{figure}[bt]
      \centering
      \begin{tikzpicture}
            \begin{axis}[
                  width=8.5cm,
                  height=5cm,
                  ymin=14, ymax=20,
                  xmin=10, xmax=64,
                  xtick = {16, 24, 32, 40, 48, 56},
                  ytick = {15, 16, 17, 18, 19},
                  xlabel={number of clusters},
                  ylabel={compression ratio},
                  grid = major,
                  legend entries={
                    equidistant,
                    k-means,
                    equidistant with tonal opt.,
                    k-means with tonal opt.},
                  legend cell align=left,
                  legend style={legend pos=north east},
                  ]
                  \addplot+[mark=none, solid, very thick, red]
                  table[x index=6, y index=4] {data/trui-1-3.tab};
                  \addplot+[mark=none, dotted, very thick, blue]
                  table[x index=6, y index=0] {data/trui-1-3.tab};
                  \addplot+[mark=none, dashed, very thick, red]
                  table[x index=6, y index=4] {data/truitest-1-3.tab};
                  \addplot+[mark=none, dashdotted, very thick, blue]
                  table[x index=6, y index=0] {data/truitest-1-3.tab};
            \end{axis}
      \end{tikzpicture}
      \caption{Performance evaluation of our quantisation schemes in terms of
        compression ratio. The solid and dotted curves denote the findings for
        an equidistant quantisation and a k-means quantisation without a tonal
        optimisation as post processing. The other two curves depict the
        findings with an additional tonal optimisation.}
      \label{fig:PlotCompRatio}
\end{figure}
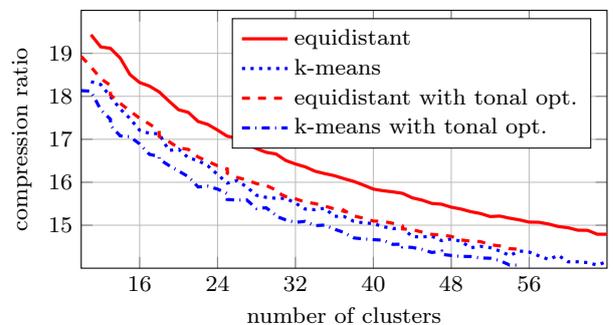

In our first experimental setup, we use the same binary mask to compare the
compression performance of equidistant and clustering-based quantisation: We
quantise the original grey values with each approach respectively and apply the
same coding techniques afterwards. Fig.~\ref{fig:Perf} shows that for low
compression ratios, equidistant and clustering-based quantisation lead to
similar results with a slight benefit for clustering. However, for higher
compression ratios, the equidistant quantisation yields clearly superior
results. In order to understand this outcome, we analyse the dependencies of the
reconstruction error and the ratio on the number of quantisation levels for both
methods in Fig.~\ref{fig:PlotInpaintErr} and Fig.~\ref{fig:PlotCompRatio}.\par
Fig.~\ref{fig:PlotInpaintErr} shows that the reconstruction error of
clustering-based quantisation fluctuates much less depending on the number of
quantised grey values than in the case of equidistant quantisation. However, it
does not offer a distinct qualitative advantage. On the contrary, for very
coarse quantisations, the equidistant quantisation can also outperform the
results of clustering in isolated cases.\par
More importantly, the non-equidistant quantisation does not only change the
reconstruction error, but also the entropy of the sequence of grey values that
need to be stored. Fig.~\ref{fig:PlotCompRatio} reveals that this yields a
consistent increase in storage costs compared to an equidistant approach.
Consequently, clustering needs to use a coarser quantisation at the cost of
reconstruction quality in order to reach the same compression ratio as an
equidistant method. Overall, this leads to the superior performance of the
equidistant quantisation.\par
Moreover, there is another factor that needs to be considered for our
evaluation: State-of-the-art compression codecs do not only carefully select the
location of known data, but also optimise the pixel values under the constraint
of the quantisation \cite{SPMEWB14,PHHW16}. Data optimisation schemes, such as
the method of Hoeltgen et al.~\cite{HSW2013} are unable to handle quantisations
of the co-domain. Nevertheless, there is a clear mutual dependency between
optimal data positions and values that should be respected in the quantisation
process. Such a so-called quantisation-aware tonal optimisation efficiently
corrects suboptimal data locations for the quantised colour values in a
post-processing scheme and leads to a much more distinct advantage of
equidistant quantisations (see Fig.~\ref{fig:Perf}). The clustering limits the
ability of the tonal optimisation to adjust the behaviour of the inpainting
algorithm locally. This is due to an inherent property of non-equidistant
quantisations: By definition, the clustering leads to sparse regions in the grey
value domain, where only few cluster centres are located while other regions are
densely populated. In isolated regions, quantisation-aware tonal optimisation
can only apply large changes for grey values while much smaller changes are
possible in dense regions. This introduces a bias to the tonal optimisation that
does not exist for equidistant quantisation: Here, each pixel value can be tuned
by the same step size between quantisation levels, and thereby allows to diverge
more significantly from the original distribution of grey values.\par
Our evaluation shows that clustering on its own is not a suitable replacement
for the simple equidistant quantisation in PDE-based compression. More complex
quantisation techniques would require a more complex integration in the full
compression pipeline, taking into account the balance of storage cost and
reconstruction quality. While such an approach to non-equidistant quantisation
would be promising, it is beyond the scope of our publication.
%
\section{Summary and Conclusions}
\label{sec:summary-conclusions}
%
Clustering makes a lot of sense for pure quantisation of existing data. Our
experiments show that we can reduce the number of colours used for PDE-based
inpainting by as much as 80\% without encountering any significant loss in the
reconstruction quality. In addition, the presented strategies are fast and easy
to implement.\par
However, the cost of the associated data and the limitations imposed on tonal
optimisation make this concept difficult to apply to compression if it is
treated as an isolated component in the compression pipeline. Our findings show
that a good quantisation with respect to the reconstruction error does not
necessarily imply an efficient encoding of the data.\par
Therefore, successful data optimisation strategies in the context of image
compression must include the encoding costs in their framework. It is not enough
to consider the reconstruction error for the quantisation.\par
Such potential new models are challenging in several ways. Not only are the
encoding costs difficult to predict, but also their optimisation is likely to be
costly. For practical applications it is however necessary to devise strategies
that are very fast since state-of-the-art data compression codecs have run times
in the millisecond range.\par
%
\bibliographystyle{spmpsci}
\bibliography{arxiv-preprint}

\begin{thebibliography}{10}
\providecommand{\url}[1]{{#1}}
\providecommand{\urlprefix}{URL }
\expandafter\ifx\csname urlstyle\endcsname\relax
  \providecommand{\doi}[1]{DOI~\discretionary{}{}{}#1}\else
  \providecommand{\doi}{DOI~\discretionary{}{}{}\begingroup
  \urlstyle{rm}\Url}\fi

\bibitem{BBBW2009}
Belhachmi, Z., Bucur, D., Burgeth, B., Weickert, J.: How to choose
  interpolation data in images.
\newblock SIAM Journal on Applied Mathematics \textbf{70}(1), 333--352 (2009)

\bibitem{B2015}
Bibiane-Schönlieb, C.: Partial Differential Equation Methods for Image
  Inpainting.
\newblock Cambridge Monographs on Applied and Computational Mathematics.
  Cambridge University Press (2015)

\bibitem{B1974a}
de~Boor, C.: Good approximation by splines with variable knots {II}.
\newblock In: G.~Watson (ed.) Conference on the Numerical Solution of
  Differential Equations, \emph{Lecture Notes in Mathematics}, vol. 363, pp.
  12--20. Springer (1974)

\bibitem{BU13}
Bourquard, A., Unser, M.: Anisotropic interpolation of sparse generalized image
  samples.
\newblock IEEE Transactions on Image Processing \textbf{22}(2), 459--472 (2013)

\bibitem{CH1974}
Calinski, R.B., Harabasz, J.: A dendrite method for cluster analysis.
\newblock Communications in Statistics \textbf{3}, 1--27 (1974)

\bibitem{CRP2014}
Chen, Y., Ranftl, R., Pock, T.: A bi-level view of inpainting - based image
  compression.
\newblock Computing Research Repository  (2014).
\newblock Available from \url{http://arxiv.org/abs/1401.4112v2}

\bibitem{DB1979}
Davies, D.L., Bouldin, D.W.: A cluster separation measure.
\newblock In: IEEE Transactions on Pattern Analysis and Machine Intelligence,
  vol. PAMI-1 2, pp. 224--227 (1979)

\bibitem{DI2003}
Demaret, L., Iske, A.: Scattered data coding in digital image compression.
\newblock In: A.~Cohen, J.L. Merrien, L.L. Schumaker (eds.) Curve and Surface
  Fitting: Saint-Malo 2002, pp. 107--117. Nashboro Press (2003)

\bibitem{DI2004}
Demaret, L., Iske, A.: Advances in digital image compression by adaptive
  thinning.
\newblock Annals of the MCFA \textbf{3}, 105--109 (2004)

\bibitem{DIK2009}
Demaret, L., Iske, A., Khachabi, W.: Contextual image compression from adaptive
  sparse data representations.
\newblock In: R.~Gribonval (ed.) Proceedings of SPARS'09 (Signal Processing
  with Adaptive Sparse Structured Representations Workshop) (2009).
\newblock Available online: \url{https://hal.inria.fr/inria-00369491}

\bibitem{D2013}
Desgraupes, B.: Clustering indices (2013).
\newblock
  \url{https://cran.r-project.org/web/packages/clusterCrit/vignettes/clusterCrit.pdf}

\bibitem{GWWB05}
Gali\'c, I., Weickert, J., Welk, M., Bruhn, A., Belyaev, A., Seidel, H.P.:
  Towards {PDE}-based image compression.
\newblock In: Variational, Geometric and Level-Set Methods in Computer Vision,
  \emph{LNCS}, vol. 3752, pp. 37--48. Springer (2005)

\bibitem{GT2001}
Gilbarg, D., Trudinger, N.: Elliptic Partial Differential Equations of Second
  Order.
\newblock Springer (2001)

\bibitem{GRBL2012}
Guerra, L., Robles, V., Bielza, C., Larranaga, P.: A comparison of clustering
  quality indices using outliers and noise.
\newblock Intelligent Data Analysis \textbf{16}, 703--715 (2012)

\bibitem{H2002}
Hamideh, H.: On the optimal knots of first degree splines.
\newblock Kuwait Journal of Science and Engineering \textbf{29}(1), 1--13
  (2002)

\bibitem{HB2016}
Hoeltgen, L., Breuß, M.: Efficient co-domain quantisation for pde-based image
  compression.
\newblock In: A.~Handlovičová, D.~Ševčovič (eds.) Proceedings of the
  Conference Algoritmy 2016, pp. 194--203. Publishing House of Slovak
  University of Technology in Bratislava (2016)

\bibitem{HHBK2017}
Hoeltgen, L., Harris, I., Breu{\ss}, M., Kleefeld, A.: Analytic existence and
  uniqueness results for pde-based image reconstruction with the laplacian.
\newblock In: F.~Lauze, Y.~Dong, A.B. Dahl (eds.) Scale Space and Variational
  Methods in Computer Vision - 6th International Conference, {SSVM} 2017,
  Kolding, Denmark, June 4-8, 2017, Proceedings, \emph{Lecture Notes in
  Computer Science}, vol. 10302, pp. 66--79. Springer (2017)

\bibitem{HSW2013}
Hoeltgen, L., Setzer, S., Weickert, J.: An optimal control approach to find
  sparse data for {L}aplace interpolation.
\newblock In: Energy Minimization Methods in Computer Vision and Pattern
  Recognition, \emph{LNCS}, vol. 8081, pp. 151--164. Springer (2013)

\bibitem{HW15}
Hoeltgen, L., Weickert, J.: Why does non-binary mask optimisation work for
  diffusion-based image compression?
\newblock In: Energy Minimization Methods in Computer Vision and Pattern
  Recognition, \emph{LNCS}, vol. 8932, pp. 85--98. Springer (2015)

\bibitem{KMHD2013}
Kreit, E., Mäthger, L.M., Hanlon, R.T., Dennis, P.B., Naik, R.R., Forsythe,
  E., Heikenfeld, J.: Biological versus electronic adaptive coloration: how can
  one inform the other?
\newblock Journal of the the Royal Society Interface \textbf{10}(78) (2013)

\bibitem{LSWL07}
Liu, D., Sun, X., Wu, F., Li, S., Zhang, Y.Q.: Image compression with
  edge-based inpainting.
\newblock IEEE Transactions on Circuits, Systems and Video Technology
  \textbf{7}(10), 1273--1286 (2007)

\bibitem{L1982}
Lloyd, S.P.: Least squares quantization in pcm.
\newblock IEEE Transactions on Information Theory \textbf{2}(28), 129--137
  (1982)

\bibitem{Ma05}
Mahoney, M.: Adaptive weighing of context models for lossless data compression.
\newblock Tech. Rep. CS-2005-16, Florida Institute of Technology, Melbourne,
  Florida (2005)

\bibitem{MHWT2012}
Mainberger, M., Hoffmann, S., Weickert, J., Tang, C.H., Johannsen, D., Neumann,
  F., Doerr, B.: Optimising spatial and tonal data for homogeneous diffusion
  inpainting.
\newblock In: A.M. Bruckstein, B.M. {Haar ter Romeny}, A.M. Bronstein, M.M.
  Bronstein (eds.) Scale Space and Variational Methods in Computer Vision,
  \emph{LNCS}, vol. 6667, pp. 26--37. Springer (2012)

\bibitem{OCBP2014}
Ochs, P., Chen, Y., Brox, T., Pock, T.: i{P}iano: Inertial proximal algorithm
  for non-convex optimization.
\newblock SIAM Journal on Imaging Sciences \textbf{7}(2), 1388--1419 (2014)

\bibitem{PHHN2015}
Peter, P., Hoffmann, S., Nedwed, F., Hoeltgen, L., Weickert, J.: From optimised
  inpainting with linear {PDE}s towards competitive image compression codecs.
\newblock In: T.~Bräunl, B.~McCane, M.~Rivers, X.~Yu (eds.) Advances in Image
  and Video Technology, \emph{LNCS}, vol. 9431, pp. 63--74. Springer (2015)

\bibitem{PHHW16}
Peter, P., Hoffmann, S., Nedwed, F., Hoeltgen, L., Weickert, J.: Evaluating the
  true potential of diffusion-based inpainting in a compression context.
\newblock Signal Processing: Image Communication \textbf{46}, 40--53 (2016)

\bibitem{PSMM15}
Peter, P., Schmaltz, C., Mach, N., Mainberger, M., Weickert, J.: Beyond pure
  quality: Progressive mode, region of interest coding and real time video
  decoding in {PDE}-based image compression.
\newblock Journal of Visual Communication and Image Representation \textbf{31},
  253--265 (2015)

\bibitem{R1987}
Rousseeuw, P.J.: Silhouettes: a graphical aid to the interpretation and
  validation of cluster analysis.
\newblock Computational and Applied Mathematics \textbf{20}, 53--65 (1987)

\bibitem{SPMEWB14}
Schmaltz, C., Peter, P., Mainberger, M., Ebel, F., Weickert, J., Bruhn, A.:
  Understanding, optimising, and extending data compression with anisotropic
  diffusion.
\newblock International Journal of Computer Vision \textbf{108}(3), 222--240
  (2014)

\bibitem{SWB2009}
Schmaltz, C., Weickert, J., Bruhn, A.: Beating the quality of {JPEG 2000} with
  anisotropic diffusion.
\newblock In: Pattern Recognition, \emph{LNCS}, vol. 5748, pp. 452--461.
  Springer (2009)

\bibitem{S1974}
Sundberg, R.: Maximum likelihood theory for incomplete data from an exponential
  family.
\newblock Scandinavian Journal of Statistics \textbf{1}(2), 49--58 (1974)

\bibitem{S1976}
Sundberg, R.: An iterative method for solution of the likelihood equations for
  incomplete data from exponential families.
\newblock Communications in Statistics – Simulation and Computation
  \textbf{5}(1), 55--64 (1976)

\bibitem{TWH2001}
Tibshirani, R., Walther, G., Hastie, T.: Estimating the number of clusters in a
  data set via the gap statistic.
\newblock J. R. Statist. Soc. B \textbf{63}(2), 411--423 (2001)

\bibitem{W1963}
Ward, J.H.: Hierarchical grouping to optimize an objective function.
\newblock Journal of the American Statistical Association \textbf{58}, 236--244
  (1963)

\bibitem{Z1910}
Zaremba, S.: Sur un probl\`{e}me mixte relatif \`{a} l'\'{e}quation de
  {L}aplace.
\newblock Bulletin de l'Acad\'{e}mie des {S}ciences de Cracovie pp. 313--344
  (1910)

\bibitem{ZA89}
Zeng, G., Ahmed, N.: A block coding technique for encoding sparse binary
  patterns.
\newblock IEEE Transactions on Acoustics, Speech, and Signal Processing
  \textbf{37}(5), 778--780 (1989)

\end{thebibliography}
\end{document}